\pdfoutput=1
\documentclass[letterpaper, 10 pt, conference]{ieeeconf}  % Comment this line out if you need a4paper

\IEEEoverridecommandlockouts                              % This command is only needed if 
\title{Visual Robot Task Planning}

\author{Chris Paxton$^{1}$, Yotam Barnoy$^{1}$, Kapil Katyal$^{1}$, Raman Arora$^{1}$, Gregory D. Hager$^{1}$
\thanks{$^{1}$Department of Computer Science, The Johns Hopkins University, Baltimore, MD, USA
{\tt\small \{cpaxton, ybarnoy1, kkatyal2, rarora8, ghager1\}@jhu.edu}}
}

\usepackage{times}
\usepackage{helvet}
\usepackage{courier}
\usepackage{latexsym}
\usepackage[english]{babel}
\usepackage{verbatim}
\usepackage{amsmath}
\usepackage{amsfonts}
\usepackage{relsize}
\usepackage{amssymb}  % assumes amsmath package installed
\usepackage{algorithm}
\usepackage{algpseudocode}
\usepackage{epstopdf}
\usepackage{listings}
\usepackage{longtable}
\usepackage{textcomp}
\usepackage{url}
\usepackage{alltt}
\usepackage[font=small]{caption}
\usepackage{subcaption}
\usepackage{graphicx}
\usepackage{color}

\newcommand{\eat}[1]{}

\usepackage{cite}
\algnewcommand{\IIf}[1]{\State\algorithmicif\ #1\ \algorithmicthen}
\algnewcommand{\EndIIf}{\unskip\ \algorithmicend\ \algorithmicif}

\begin{document}

\maketitle
\thispagestyle{empty}
\pagestyle{empty}

%%%%%%%%%%%%%%%%%%%%%%%%%%%%%%%%%%%%%%%%%%%%%%%%%%%%%%%%%%%%%%%%%%%%%%%%%%%%%%%%

\begin{abstract}
Prospection, the act of predicting the consequences of many possible futures,
is intrinsic to human planning and action, and may even be at the root of consciousness.
Surprisingly, this idea has been explored comparatively little in robotics.
In this work, we propose a neural network architecture and associated planning
algorithm that (1) learns a representation of the world useful for generating
prospective futures after the application of high-level actions,
(2) uses this generative model to simulate the result of sequences of
high-level actions in a variety of environments, and (3) uses this same
representation to evaluate these actions and perform tree search to find a
sequence of high-level actions in a new environment.
Models are trained via imitation learning on a variety of domains, including
navigation, pick-and-place, and a surgical robotics task.
Our approach allows us to visualize intermediate motion goals and learn to plan
complex activity from visual information.

\end{abstract}

%%%%%%%%%%%%%%%%%%%%%%%%%%%%%%%%%%%%%%%%%%%%%%%%%%%%%%%%%%%%%%%%%%%%%%%%%%%%%%%%

\IEEEpeerreviewmaketitle

\section{Introduction}
%How can we allow robots to plan as humans do?
Humans are masters at solving problems they have never encountered before.
When attempting to solve a difficult problem, we are able to build a good abstract
models and to picture what effects our actions will have.
Some say this act --- the act of prospection --- is the essence of
true intelligence~\cite{Seligman2013}. If we want robots that can plan
and act in general purpose situations just as humans do, this ability would appear
crucial.

As an example, consider the task of stacking a series of colored blocks in a
particular pattern, as explored in prior work~\cite{xu2017neural}.
A traditional planner would view this as a sequence of high-level actions, such
as \texttt{pickup(block)}, \texttt{place(block,on\_block)}, and so on. The
planner will then decide which object gets picked up and in which order.
Such tasks are often described using a formal language such as the Planning
Domain Description Language (PDDL)~\cite{ghallab1998pddl}.
%Success and constraints for these actions would need to be defined, goal conditions, cost functions, et cetera. \RA{Revise the previous sentence.}
To execute such a task on a robot, specific goal conditions and cost functions must be defined,
and the preconditions and effects of each action must be specified.
This is a time consuming manual undertaking~\cite{beetz2012cognition}.
Humans, on the other hand, do not require that all of this information to be given
to them beforehand. We can learn models of task structure purely from
observation or demonstration. We work directly with high dimensional data
gathered by our senses, such as images and haptic feedback, and can reason over
complex paths without being given an explicit model or structure.

%\begin{figure}[bt]
%        \centering
%        \includegraphics[width=\columnwidth]{validation_v2}
%        \caption{Predicting results of high-level actions performed during a simple stacking task including an obstacle that must be
%        avoided.}
%                %The robot must decide which blocks to pick up and move, and which block to
%                %put them on, taking into account its workspace and the obstacle. Given two
%    %high-level actions like \texttt{approach(red\_block)},
%    %\texttt{grasp(red\_block)}, we can generate corresponding goal states.}
%        \label{fig:blocks-example}
%\end{figure}
\begin{figure}[bt!]
  \centering
  \includegraphics[width=0.8\columnwidth]{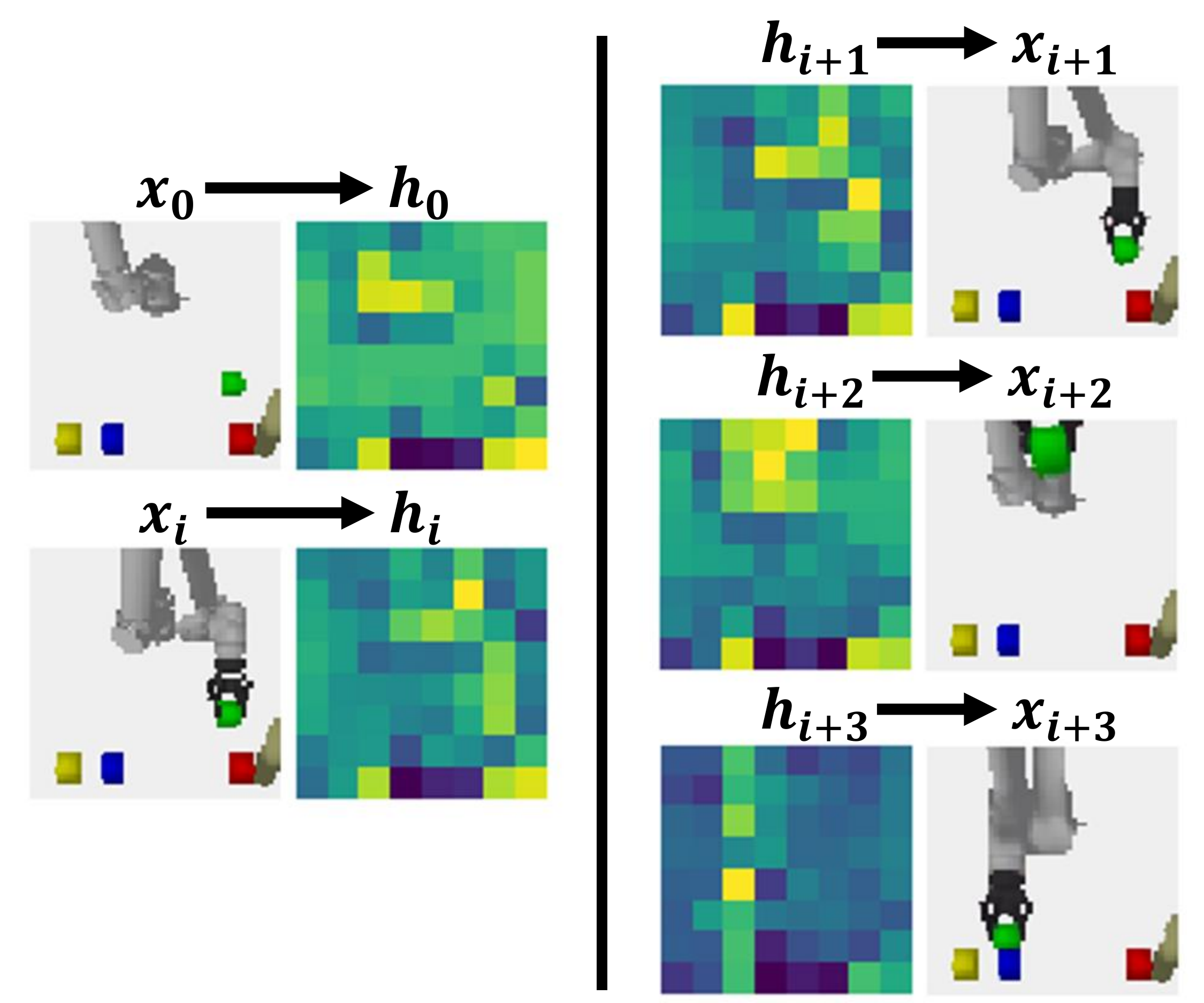}
  \caption{Example of our algorithm using learned policies to predict a good
  sequence of actions. Left: initial observation $x_0$ and current observation
$x_i$, plus corresponding encodings $h_0$ and $h_i$.
Right: predicted results of three sequential high level actions.}
  \label{fig:good-example}
  \vskip -0.5cm
\end{figure}
\begin{figure}[bt]
    \centering
    \includegraphics[width=\columnwidth]{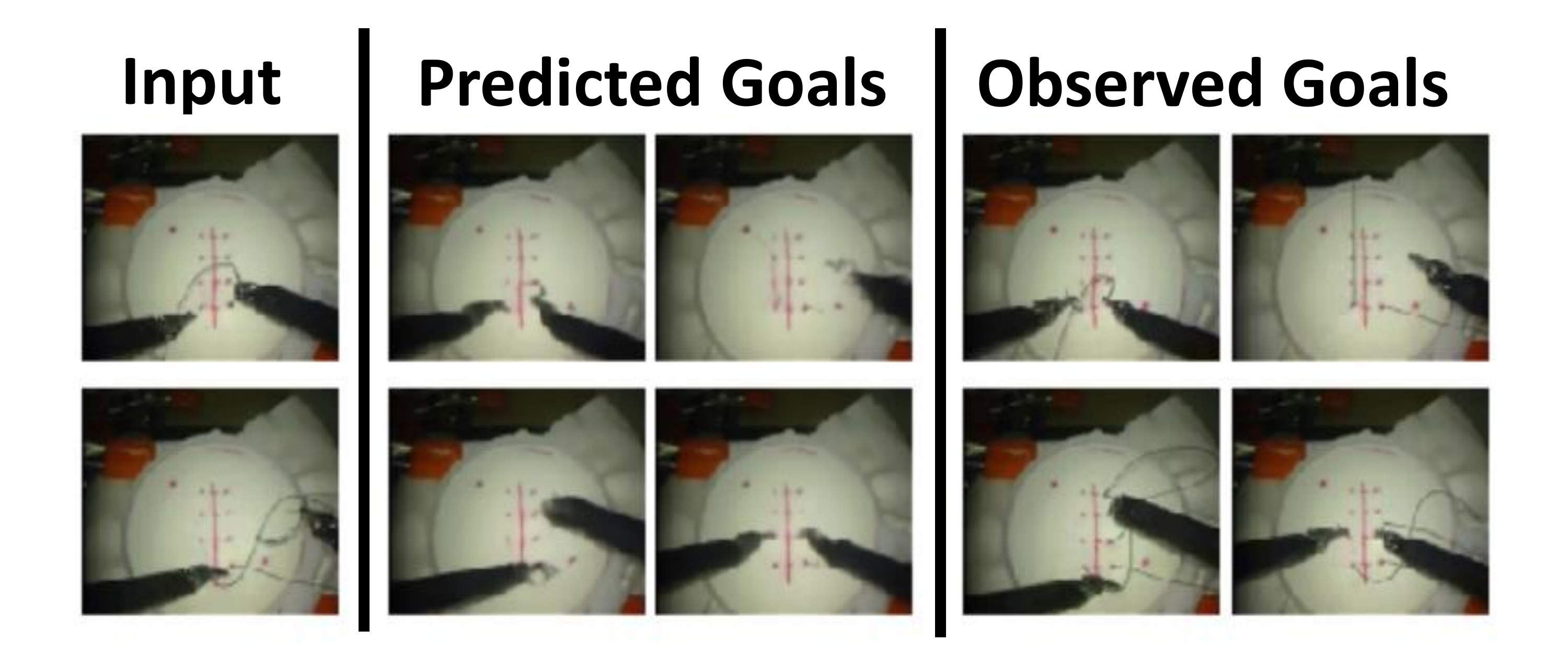}
    \caption{Predicting the next step during a suturing task based on labeled
    surgical data. Predictions clearly show the next position of the arms.}
    \label{fig:suturing}
  \vskip -0.5cm
\end{figure}

%In fact, one of the key problems when deploying robots in new environments is
%that of specifying the problem definition.
%For this reason, much work in robot task and motion planning (TAMP) is focused in
%relatively narrow areas such as pick-and-place tasks or coverage
%%(e.g.,~\cite{lagriffoul2014efficiently,toussaint2015logic}).
%In well-understood domains with available purpose-built perception, where task
%goals and actions are well specified, great things can be accomplished; see, for
%example, the RoboEarth project~\cite{di2013roboearth}.
%As a result, there is growing interest in grounding the definitions of these
%problems using machine learning~\cite{ahmadzadeh2015learning,PaxtonRHK17}.

% -------------------------------------------------------------------------
Ideally, we would learn representations that could be used for all aspects of
the planning problem, that also happen to be human-interpretable.
For example, deep generative models such as conditional generative adverserial networks 
(cGANs) allow us to
generate realistic, interpretable future scenes~\cite{DBLP:journals/corr/IsolaZZE16}.
In addition, a recent line of work in robotics focuses on making structured predictions to
inform planning~\cite{finn2016deep,finn2016unsupervised}:
So far, however, these approaches focus on making relatively short-term predictions, and
do not take into account high-level variation in how a task can be performed.
%In general, deep policy learning has proven successful at learning well-scoped,
%short horizon robotic
%tasks~\cite{levine2016end,sung2016robobarista,finn2016unsupervised}.
Recent work on one-shot deep imitation learning can produce general-purpose models
for various tasks, but relies on a task solution from a human expert, and
does not generate prospective future plans that can be evaluated for reliable
performance in new environments~\cite{xu2017neural}.
These approaches are very data intensive: Xu et al.~\cite{xu2017neural} used
100,000 demonstrations for their block-stacking task.
%Weighting Finite State Transductions with Neural Context~\cite{}
%Finn et al. proposed deep autoencoder models for robot control~\cite{finn2016deep,finn2016unsupervised}, and~\cite{byravan2017se3}

We propose a supervised model that learns high-level task
structure from imperfect demonstrations. Our approach then generates interpretable task plans by predicting
and evaluating a sequence of high-level actions, as shown in
Fig.~\ref{fig:good-example}.
Our approach learns a pair of functions $f_{enc}$ and $f_{dec}$
that map into and out of a lower-dimensional hidden space $\mathcal{H}$, and then learns a set
of other functions that operate on values in this space.
Models are learned from labeled training data containing both successes and
failures, are not reliant on a large number of good expert examples, and
work in a number of domains including navigation, pick-and-place, and robotic
suturing (Fig.~\ref{fig:suturing}).
We also describe a planning algorithm that uses these results to simulate a set of possible
futures and choose the best sequence of high-level actions to execute, resulting
in realistic explorations of possible futures.

To summarize, our contributions are:
\begin{itemize}
  \item A network architecture and training methodology for learning a deep
    representation of a planning task.
  \item An algorithm to employ this planning task to generate and evaluate
    sequences of high-level actions.
  \item Experiments demonstrating the model architecture and algorithm on
    multiple datasets.
\end{itemize}
Datasets and source code will be made available upon publication.

\section{Related Work}\label{sec:background}

\subsubsection*{Motion Planning}
In robotics, effective TAMP approaches have been developed for solving complex
problems involving spatial reasoning~\cite{toussaint2015logic}.
A subset of planners focused on Partially Observed Markov Decision Process
extend this capability into uncertain worlds; examples include
DeSPOT, which allows manipulation of objects in
cluttered and challenging scenes~\cite{li2016act}.
These methods rely on a large amount of built-in knowledge about the world,
however, including object dynamics and grasp locations.

A growing number of works have explored the integration of
planning and deep neural networks.
For example, QMDP-nets embed learning into a planner
using a combination of a filter network and a value function approximator
network~\cite{karkus2017qmdp}.
Similarly, value iteration networks embed a
differentiable version of a planning algorithm (value iteration) into a neural
network, which can then learn navigation tasks~\cite{tamar2016value}.
Vezhnevets proposed to generate plans as sequences of
actions~\cite{vezhnevets2016strategic}.
Other prior work employed Monte Carlo Tree Search (MCTS) together
with a set of learned action and control policies for task and motion
planning~\cite{PaxtonRHK17}, but did not incorporate predictions.

%\textbf{In this work, we examine the problem of learning representations for high-level predictive models for task planning.}
Prediction is intrinsic to planning in a complex world.
While most robotic motion planners assume a simple causal model of the world,
recent work has examined learning predictive models.
%In \cite{ondruska2016deep}, the authors fit a model to predict the true state of an
%occluded world as it evolves over time.
Lotter et al.~\cite{lotter2016deep} propose PredNet as
a way of predicting sequences of images from sensor data, with the goal of
predicting the future, and Finn et al.~\cite{finn2016unsupervised} use unsupervised learning of
predictive visual models to push objects around in a plane.
However, to the best of our knowledge,
ours is the first work to use prospection for task planning.

%\subsubsection*{Policy Learning}
%The \emph{options framework} provides a way to think of MDPs as a set of many high level ``options,'' each active over a certain window~\cite{suttonPS99}.
%The authors in~\cite{andreas2016modular} propose a method that uses curriculum learning with a set of  predefined ``policy sketches'' to solve planning problems in the options framework.
%Another recent approach is FeUdal networks, in which a ``manager'' network sets goals
%for lower level ``worker'' networks~\cite{vezhnevets2017feudal}.

\subsubsection*{Learning Generative Models.}
%Such a prediction system must be able to deal with a stochastic world.
GANs are widely considered
the state of the art in learned image generation~\cite{chen2017photographic,arjovsky2017wasserstein},
though they are far from the only option.
The Wasserstein GAN is of particular note as an improvement over other GAN
architectures~\cite{arjovsky2017wasserstein}.
Isola et al. proposed the PatchGAN, which uses an average adversarial loss over ``patches'' of the image,
together with an L1 loss on the images as a way of training
conditional GANs to produce one image from another~\cite{DBLP:journals/corr/IsolaZZE16}.

Prior work has examined several ways of generating multiple realistic predictions~\cite{rupprecht2017learning,chen2017photographic,ghadirzadeh2017deep}.
%A key challenged is the design of an appropriate loss function, particularly in the case of many possible outcomes.
The authors in~\cite{ho2016generative} demonstrated the advantage
of applying adversarial methods to imitation learning.
More recently,~\cite{ghadirzadeh2017deep} proposed to learn a deep predictive
network that uses a stochastic policy over goals for manipulation tasks, but
without the goal of additionally predicting the future world state.

\subsubsection*{Learning Representations for Planning}
Sung et al.~\cite{sung2016robobarista} learn a deep multimodal embedding for a variety of tasks.
This representation allows them to adapt to appliances with
different interfaces while reasoning over trajectories,
natural language descriptions, and point clouds for a given task.
%They do this by embedding the trajectory, natural language description, and point cloud associated with a task in a latent space.
Finn et al.~\cite{finn2016deep} learn a deep autoencoder as a set of convolutional blocks followed by a spatial softmax; they found that this representation was useful for reinforcement learning.
Recently, Higgins et al.~\cite{higgins2017darla} proposed DARLA, the DisentAngled Representation
Learning Agent, which learns useful representations for tasks that enable
generalization to new environments. Interpretability and predicting far into the
future were not goals of these approaches.

\section{Approach}\label{sec:approach}

We define a planning problem with continuous states $x \in \mathcal{X}$,
where $x$ contains observed information about the
world such as a camera image.
We augment this state with high-level actions $a \in A$ that describe the task structure.
We also assume that the hidden world state $h \in \mathcal{H}$ should encode
both task information (such as goals) and the underlying ground truth
input from the various sensors.
For example, in the block-stacking task in Fig.~\ref{fig:good-example}, $h$
encodes the positions of the four blocks, the obstacle and the
configuration of the arm.

\begin{figure*}[bht!]
  \centering
  \includegraphics[width=1.6\columnwidth]{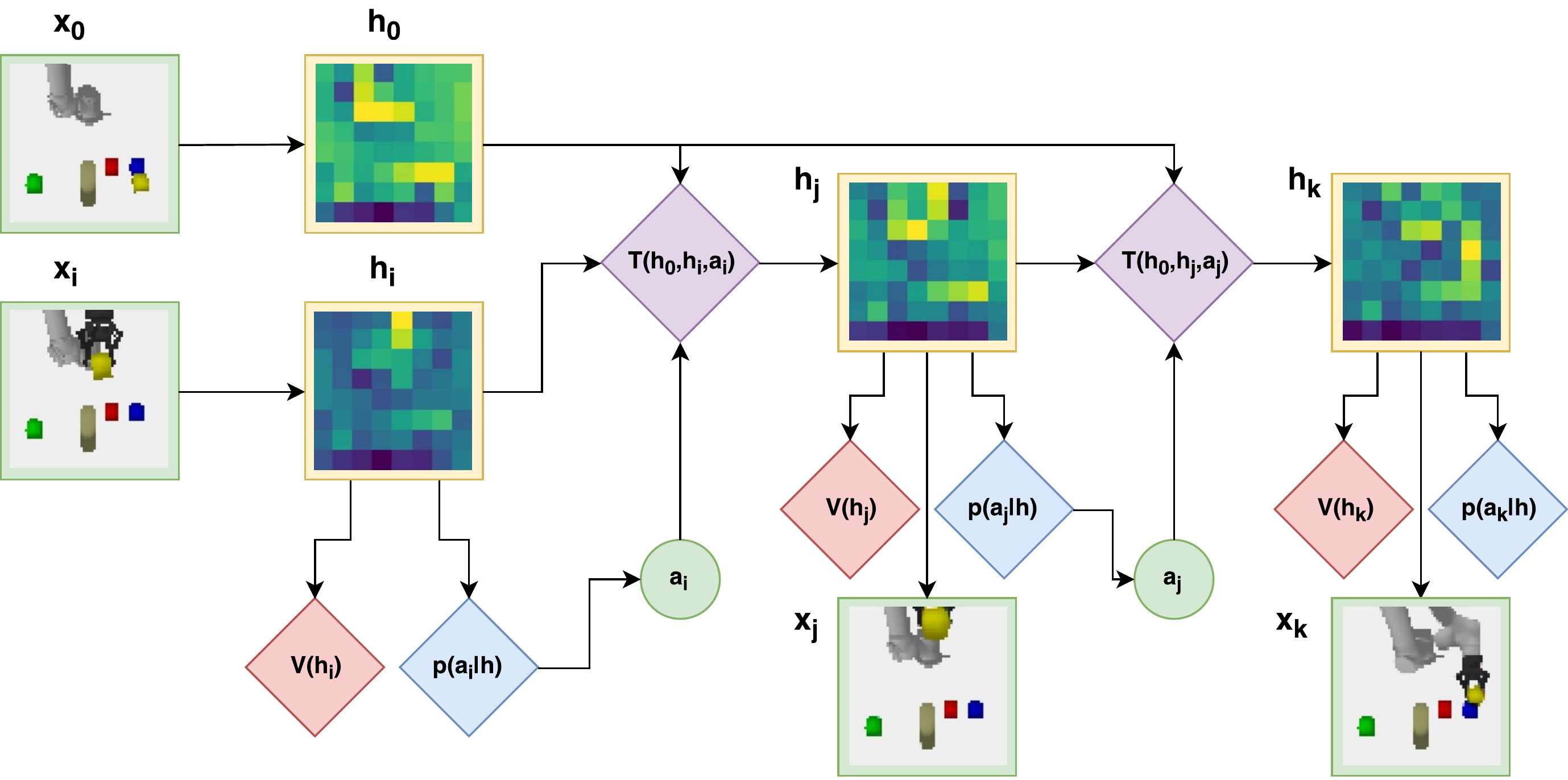}
  \caption{Overview of the prediction network for visual task planning. We learn
  $f_{enc}(x)$, $f_{dec}(x)$, and $T(h, a)$ to be able to predict and visualize
results of high-level actions.}
  \label{fig:vtp}
  \vskip -0.5cm
\end{figure*}

%Our goal is to learn models grounding this problem as an MDP in the options framework.%, so that at run time we can generate intelligent, comprehensible task plans.
Our objective is to learn a set of models representing the necessary components of
this planning problem, but acting in this latent space $\mathcal{H}$.
In other words, given a particular action $a$ and an observed state $x$, we want to be able to predict
both an end state $x'$ and the optimal sequence of actions $a \in A^*$
necessary to take us there.
We specifically propose that there are three components of this prediction function:
\begin{itemize}
  \item[1.] $f_{enc}(x)\rightarrow h \in \mathcal{H}$, a learned encoder function maps observations and descriptions to the hidden state.
	\item [2.] $f_{dec}(h) \rightarrow (x)$, a decoder function that maps from the hidden state of the world to the observation space.
  \item [3.] $T(h, a) \rightarrow h' \in \mathcal{H}$, the $i$-th learned world state transformation function, which maps to different positions in the space of possible hidden world states.
\end{itemize}

Specifically, we will first learn an action subgoal prediction function, which is a
mapping $f_{dec}(T(f_{enc}(x),a)) \rightarrow (x')$.
In practice, we include the hidden state of the first world observation as well
in our transform function, in order to capture any information about the world
that may be occluded. This gives the transform function the form:
\[
  T(h_0, h, a) \rightarrow h' \in \mathcal{H}
\]

We assume that the hidden state $h$ contains all the necessary information about
the world to make high level decisions as long as this $h_0$ is available to
capture change over time. As
such, we learn additional functions representing the
value of a given hidden state, the predicted value of actions moving forward
from each hidden state, and connectivity between hidden states.
Given these components, we can appropriately represent the task as a tree search problem.

\subsection{Model Architecture}\label{sec:model}

Fig.~\ref{fig:vtp} shows the architecture for visual task planning. Inputs are
two images $x_0$ and $x_i$: the initial frame when the planning problem was
first posed, and the current frame. We include $x_0$ and $h_0$ to capture
occluded objects and changes over time from the beginning of the task.
In Fig.~\ref{fig:vtp}, hidden states $h_0, h_i, h_j, h_k$ are represented by
averaging across channels.

% https://www.draw.io/#G1aMt6evXo5FQ2q8A4RN12JZnB7D6xN9Cx
\begin{figure*}[bth!]
  \centering
  \includegraphics[width=1.8\columnwidth]{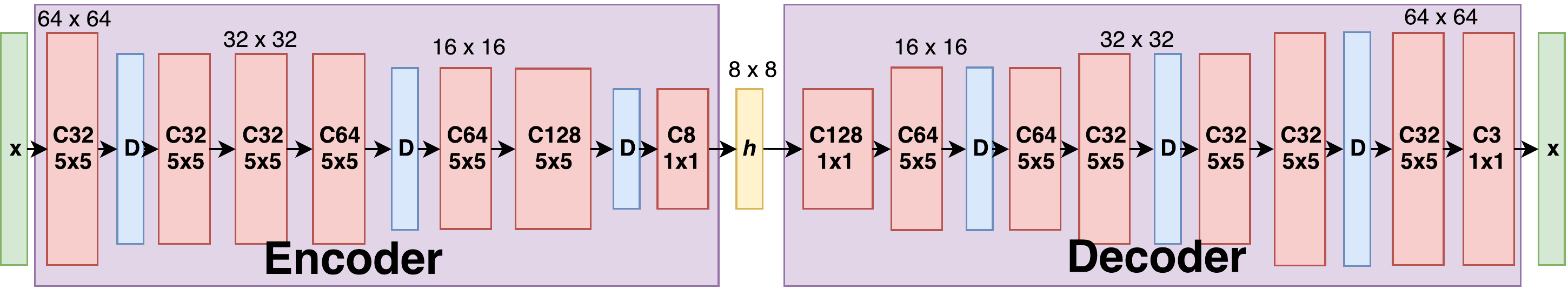}
  \caption{Encoder-decoder architecture used for learning a transform into and
  out of the hidden space $h$.}
  \label{fig:enc-dec}
\end{figure*}

\textbf{Encoder and Decoder.} The first training step find the transformations into and out of the
learned hidden space $\mathcal{H}$.
Specifically, we train $f_{enc}$ and $f_{dec}$ using the encoder-decoder
architecture shown in Fig.~\ref{fig:enc-dec}.
Convolutional blocks are indicated with $Ck$, where $k$ is the number of filters.
Most of our layers are $5 \times 5$ convolutions, although we used a
$7 \times 7$  convolution on the first layer and we use $1 \times 1$
convolutions to project into and out of the hidden space. Each convolution is
followed by an instance normalization and a ReLU activation. 
Stride 2 convolutions and transpose convolutions are then used to increase or
decrease image size after each block.
The final projection into the hidden state has a sigmoid activation in place of
ReLU and is not paired with a normalization layer.
In most of our examples, this hidden space is scaled down to an $8 \times 8
\times 8$ space.

Both dropout and normalization played an important role in
learning fast, accurate models for encoding the hidden state, but we use the
instance norm instead of the more common batch normalization in order to avoid
issues with dropout.
After every block,
we add a dropout layer $D$, with dropout initially set to 10\%. 
Instance normalization has been found useful for image generation in the
past~\cite{ulyanov2017improved}.
We do not apply dropout at test time except with GANs.

\begin{figure*}[bt!]
  \centering
  \includegraphics[width=1.8\columnwidth]{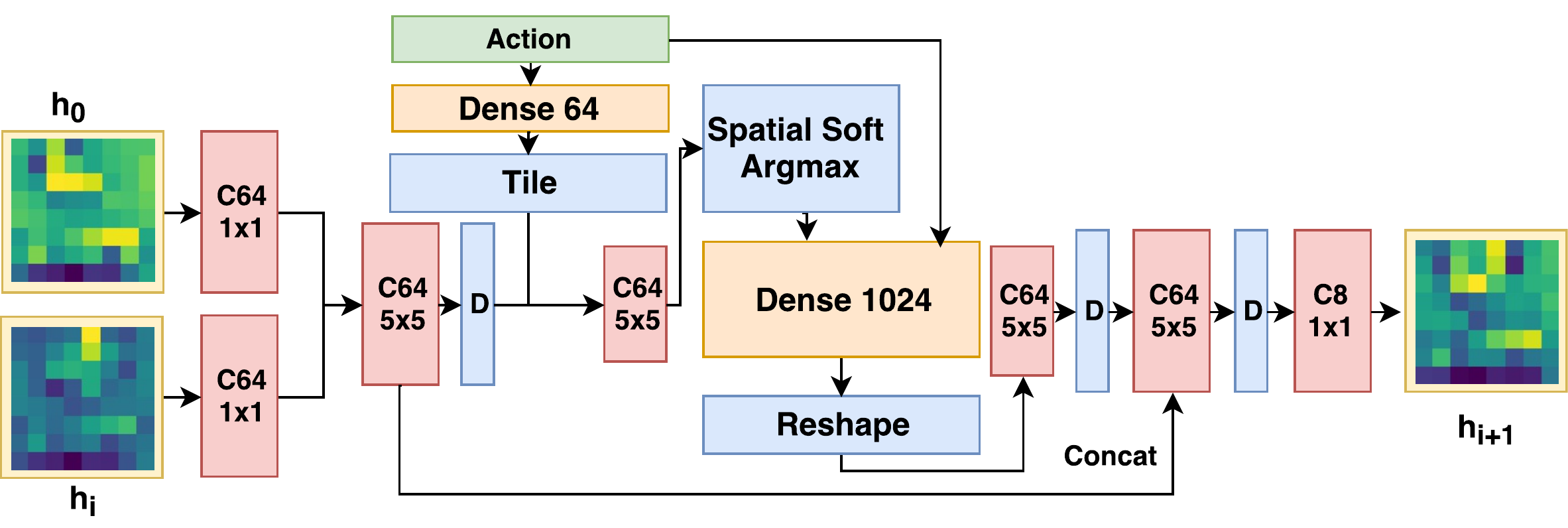}
  \caption{Architecture of the transform function $T(h_0, h, a)$ for computing transformations to
  an action subgoal in the learned hidden space.}
  \label{fig:transform}
  \vskip -0.5cm
\end{figure*}

\textbf{Transform function.} $T(h_0, h, a)$ computes the most likely next hidden
state. This function was designed to combine information about the action and
two observed states, and to compute global
information over the entire hidden space. We use the spatial soft argmax previously employed by Finn
and Levine.~\cite{levine2016end,finn2016deep} and Ghadirzadeh et
al.~\cite{ghadirzadeh2017deep} to compute a set of keypoints, which we then
concatenate with a high-level action label and use to predict a new image. This
is sufficient to capture the next action with a good deal of fidelity (see
Sec.~\ref{sec:results-architecture}), but to
capture background details we add a skip connection across this spatial soft
argmax bottleneck.

Fig.~\ref{fig:transform} shows the complete transform block as used in the block
stacking and navigation case studies described below. For the suturing case
study, with a larger input and hidden space, we add an extra set of size 64
convolutions to each side of the architecture and a corresponding skip
connection, but it is otherwise the same.

Each dense layer is followed by a ReLU nonlinearity. The final projection into
the hidden state also has a sigmoid activation and no instance normalization
layer, as in the encoder.

%In addition, we learn two functions $V(h)$ and $Q(a | h)$ which predict the
%value, or expected reward to come, as observed at hidden state $h$ and the
%probabilities of possible high level actions observed from hidden state $h$
%respectively.
%The value function $V(h)$ is trained using the full data set, and can be learned end-to-end with the encoder-transform-decoder architecture. The value function operates on the current hidden state $h$ and returns the expected reward-to-go -- i.e. whether or not we expect to see a possible success or failure farther on in the task if we continue down this route.
%The action prior $p(a | h)$ is learned on successful training data only. The goal of this function is to tell us which actions to explore first, when performing a tree search over different possible futures. This is useful because performing a tree expansion is a relatively expensive process.
%We also provide this action prior as an additional input to the transform block.
\textbf{Value functions.}
$V(h)$ computes the value of a particular hidden state $h$ as the
probability the task will be successful from that point onwards, and $Q(h_0, h,
a, a')$ predicts the probability that taking action $a'$ from the tree search node
$(h_0, h, a)$ will be successful.
These are trained based on $\left\{0, 1\right\}$ labels indicating observed task
success and observed failures.
We also train the function $f(h_0, h, a)$ which predicts whether
or not an action $a$ successfully finished.

\textbf{Structure prior.} Value functions do not necessarily indicate what
happens if there are no feasible actions from a particular state. To handle
this, we learn the permissability function $p(a' | h_0, h, a)$, which states
that it is possible for $a'$ to follow $a$, but does not state whether or not
$a'$ will succeed.

These last four models are trained on supervised data, but
without the instance normalization in $f_{emc}$, $f_{dec}$, and $T$,
as we saw this hurt performance. $Q$, $p$, and $f$ were trained with two $1
\times 1$ convolutions on $h$ and $h_0$, then $a$
concatenated, followed by $C64-C64-FC256-FC128$, where $FCk$ is a fully
connected layer with $k$ neurons. The value function $V(h)$ was a convolutional
neural net of the form $C32-C64-C128-FC128$.

%\subsubsection*{Action and Pose Prediction}
%When predicting the goals of specific actions via direct regression, we use a
%model approach informed by prior models for grasp prediction and imitation
%learning~\cite{redmon2015real}, with two large dense layers at the end of the
%network to compute poses.
%We found that this approach had the best performance in practice.

\subsection{Learning}\label{sec:learning-approach}

We train our predictor directly on supervised pairs containing the state $x' = f_{dec}(T(f_{enc}(x),
a))$ resulting from action $a$.
First, we considered a simple L1 loss on the output images.
However, this might not capture all details of complex scenes, so
we also train with an augmented loss which encourages correct classification of the
resulting image. This approach is in some ways similar to that used
by~\cite{chen2017photographic}, in which the authors predict images while
minimizing distance in a feature space trained on a classification problem.
Here, we use a combination of an L1 term and a term maximizing the
cross-entropy loss on classification of the given image, which we refer to as
the L1+$\lambda C$ loss in the following, where $\lambda$ is some weight.
Finally, we explored using two different GAN losses:
%The Wasserstein GAN~\cite{arjovsky2017wasserstein} loss to train the conditional image
%prediction model.
%The Wasserstein GAN is a variant of the Genthat has better performance in a variety of
%domains~\cite{arjovsky2017wasserstein}. We also compare to the PatchGAN
the Wasserstein GAN~\cite{arjovsky2017wasserstein} and the 
pix2pix GAN from Isola et al.~\cite{DBLP:journals/corr/IsolaZZE16}.

First we train the goal classifier $C(x)$ on labeled training data for use in
testing and in training our augmented loss.
Next we train the encoder and decoder structure $f_{enc}(x)$ and $f_{dec}(x)$,
which provide our mapping in and out of the hidden world state.
Finally we train the transform function $T(h,a)$ and the evaluation functions
used in our planning algorithm.

\textbf{Transform Training}. To encourage the model to make predictions that
remain consistent over time, we link two consecutive transforms with shared weights, and
train on the sum of the L1 loss from both images, with the optional classifier
loss term applied to the second image. The full training loss given ground truth predictions $\hat{x}_{1}$, $\hat{x}_{2}$ is then:
\[
  {\mathcal L}(x_{1}, x_{2}) = \|\hat{x}_{1} - x_{1}\|_1 +
  \|\hat{x}_{2} - x_{2}\|_1 + \lambda C(x_{2})
\]

\textbf{Implementation.} All models were implemented in Keras and trained using the Adam optimizer,
except for the Wasserstein GAN, which we trained with RMSProp as per prior work~\cite{arjovsky2017wasserstein}.
We performed
multiple experiments to set the learning and dropout rates in our models, and
selected a relatively low learning rate of $1e-4$ and dropout rate of $0.1$,
which strikes a balance between regularization and crisp per-pixel predictions
when learning the hidden state.

\subsection{Visual Planning with Learned Representations}

We use these models together %as a part of a simple tree search algorithm based on 
with Monte Carlo Tree Search (MCTS) in order to find a sequence of actions that we
believe will be successful in the new environment.
The general idea is that we run a loop where we repeatedly sample a possible
action $a'$ according to the learned function $Q(h_0, h, a, a')$ and use this action to simulate the effects
of that high-level action $T(h_0, h, a')$.
We can then execute the sequence of learned or provided black box
policies to complete the motion on the robot.

We propose a variant of MCTS as a general way of exploring
the tree over possible actions~\cite{PaxtonRHK17}.
We represent each node in
the tree by a unique instance of a high-level action ($\emptyset$ for the root).
The full algorithm is described in Alg.~\ref{alg:learned}.

\begin{algorithm}[bt!]
  \caption{Algorithm for visual task planning with a learned state representation.}
\label{alg:learned}
\begin{algorithmic}
  \small
  \State Given: max depth $d$, initial state $x_0$, current state $x$, number of
  samples $N_{samples}$
  \State $h = f_{enc}(x)$, $h_0 = h$
  \For {$i \in N_{samples}$}
  \State \Call{Explore}{$h_0$,$h$,$\emptyset$,0,$d$}
  \EndFor
  \Function{Explore}{$h_0$,$h$,$a$,$i$,$d$}
  %\State $a' \sim \pi_A(h, a)$ \Comment Sample next high-level action

  \State $v_i =$ \Call{Evaluate}{$h_0$, $h$, $a$}
  \IIf{$i \ge d$ or $v_i < v_{failed}$} \Return $v_i$ \EndIIf
  \State $a' =$ \Call{Sample}{$h_0$, $h$, $a$}
  %\If{$N(a, a') = 0$}
  \State $h' = T(h_0, h, a')$ %\Comment Compute result of $a'$
  %\Else
    %\State $h' =$ \Call{Lookup}{$a$,$a'$}
  %\EndIf
  \State $v' =$ \Call{Explore}{$h_0$,$h'$,$a'$,$i+1$,$d$}
  \State \Call{Update}{$a$, $a'$, $v'$}
  \State \Return $v_i \cdot v'$
  \EndFunction
\end{algorithmic}
\end{algorithm}

The {\sc{Evaluate}} function sets $v_i = V(h)$, but also checks the validity of the
chosen action and determines which actions can be sampled.
We also compute $f(h_0, h, a)$ to determine if the robot would succesfuuly complete the action with
some confidence $c_{done}$. If not this is considered a failure ($v_i = 0$).
If $v' < v_{failed}$, we will halt exploration.

The {\sc{Sample}} function greedily chooses the next
action $a'$ to pursue according to a score $v(a,a')$:
\[
  v(a,a') = \frac{c Q(h_0, h, a, a')}{N(a,a')} + v^*(a,a')
\]
\noindent where $Q$ is the learned action-value function, $N(a, a')$ is the
number of times $a'$ was visited from $a$, and $v^*(a, a')$ is the best observed
result from taking $a'$. We set $c = 10$ to encourage exploration to actions
that are expected to be good. The {\sc{Update}} function is responsible for
incrementing $N(a,a')$.
Sampled actions $a'$ are rejected if we predict $a'$ is not reachable from its
parent.

\section{Experimental Setup}\label{sec:experiment}

We applied the proposed method to both a simple navigation task using a
simulated Husky robot, and to a UR5 block-stacking task.
\footnote{Source code for all examples will be made available after publication.}

In all examples, we follow a simple process for collecting data. First, we generate a random world configuration, determining where objects will be placed in a given scene. The robot is initialized in a random pose as well. We automatically build a task model that defines a set of control laws and termination conditions, which are used to generate the robot motions in the set of training data.
Legal paths through this task model include any which meet the high-level task specification, but may still violate constraints (due to collisions or errors caused by stochastic execution).

We include both positive and negative examples in our training data.
Training was performed with Keras~\cite{chollet2015keras} and Tensorflow 1.5 for
45,000 iterations on an NVidia Titan Xp GPU, with a batch size of 64. Training
took roughly 200 ms per batch.

\subsection{Robot Navigation}

\begin{figure}[bt]
        \centering
        \includegraphics[width=\columnwidth]{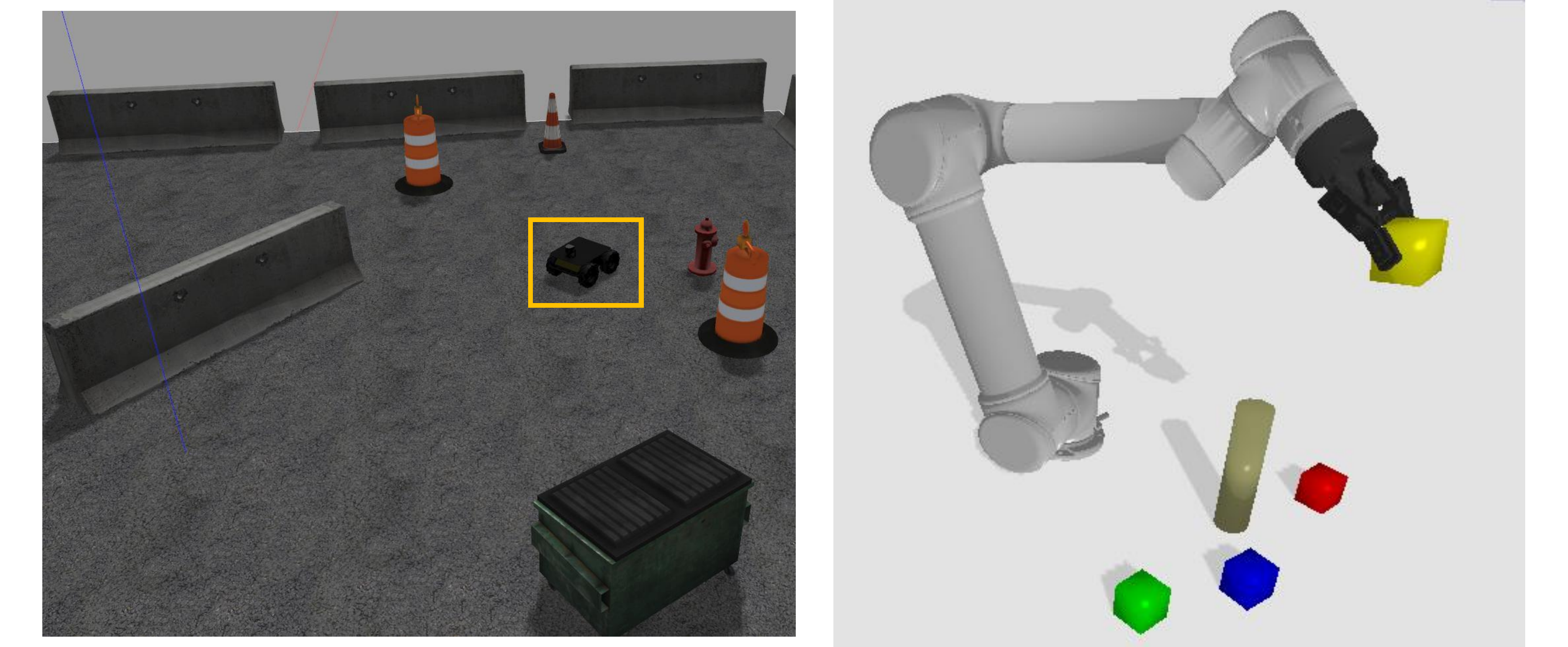}
        \caption{Simulation experiments. Left: Husky navigation task. The robot
        is highlighted. Right: UR5 block-stacking task with obstacle avoidance.}
        \label{fig:sim}
        \vskip -0.5cm
\end{figure}

In the navigation task, we modeled a Husky robot moving through a construction
site environment to
investigate one of four objects: a barrel, a barricade, a construction pylon or
a block, as shown in Fig.~\ref{fig:sim}(left).
The goal was to find a path between different objects, so that it takes less then
ten seconds to move between any two objects.
Here, $x$ was a 64x64 RGB image that provides an aerial view of the
environment. Data was collected using a Gazebo simulation of the robot
navigating to a randomly-chosen sequence of objects.
We collected 208 trials, of which 128 were failures.

\subsection{Simulated Block Stacking}

To analyze our ability to predict goals for task planning, we learn in a more elaborate environment.
In the block stacking task, the robot needed to pick up a colored block and place it on top of any other colored block.
The robot succeeds if it manages to stack any two blocks on top of one another
and failed immediately if either it touches this obstacle or if at the end of 30
seconds the task has not been achieved. Training was performed on a relatively
small number of examples: we used 6020 trials, of which 2991 were successful and
3029 were failures. %The task is shown in Fig.~\ref{fig:sim}.
%In this case, we represent the robot in terms of its 6DOF end effector pose $x_{ee}$,
%encoded as the concatenation of the position $(x, y, z)$ and the
%roll-pitch-yaw$(\omega_r, \omega_p, \omega_y)$, such that $x_{ee} = \left\{x, y,
%z, \omega_r, \omega_p, \omega_y\right\}$. The state of the gripper was expressed
%as a single variable $x_g \in (0, 1)$. In addition,
The state $x$ is a $64 \times 64$ RGB image of the scene from a fixed external camera.
%Fig.~\ref{fig:blocks-example} shows the results of this process.
%Scenes included four blocks and the obstacle in addition to the robot, but lack any other objects or background clutter.

We provided a set of non-optimal expert policies and randomly sampled a set of
actions. This task was fairly complicated, with a total of 36 possible actions
divided between two sub-tasks. Separate high-level actions were provided for
aligning the gripper with an object, moving towards a grasp, closing the
gripper, lifting an object, aligning a block with another block below it,
stacking the currently held block on another, opening the gripper, and returning
to the home position. These were provided for each of four colored blocks the
robot could manipulate: the actions were either parameterized by the block to
pick up (the first four steps) or the block to stack on top of (the last four
steps).
%The entire task model can be seen in Fig.~\ref{fig:task-model-stacking}.
%Task performances were sampled at random.

Each performance was labeled a failure
if either (a) it took more than 30 seconds, (b) there was a collision with the
obstacle, or (c) the robot moved out of the workspace for any reason.
The simulation was implemented in PyBullet, and was designed to be stochastic
and unpredictable. At times the robot would drop the currently-held block, or it
would fail to accurately place the block held in its hands. There was also some noise
in the simulated images.

\subsection{Surgical Robot Image Prediction}

Next, we explored our ability to predict the goal of the next motion on a
real-world surgical robot problem.
Minimally invasive surgery is a highly
skilled task that requires a great deal of training; our image prediction
approach could allow novice users some insight into what an expert might do in
their situation.
There is a growing amount of surgical robot video available, and a
growing body of work seeks to capitalize on this to improve video
prediction~\cite{vedula2016analysis}.
We used a subset of the JIGSAWS dataset to train a
variant of our Visual Task Planning models on a labeled suturing task in order
to predict the results of certain motions.

The JIGSAWS dateset consists of stereo video frames, with each pair labeled
as belonging to one of 15 possible gestures.
For our task, We used only the left frames of the video stream.
The dataset contains the 3 tasks of suturing, know tying, and needle passing,
with each task consisting of a subset of gestures.
We reduced the image dimensions from $640\times480$ to a more reasonable $96\times128$.
For this application, we used a slightly larger $12\times16\times8$ hidden
representation.
Fig.~\ref{fig:pretrain-suturing} shows examples of the pretrained $f_{enc}$ and
$f_{dec}$ functions.

\section{Results}

Our models are able to generate realistic predictions of possible futures for
several different tasks, and can use these predictions to make intelligent
decisions about how they should move to solve planning problems.
See Fig.~\ref{fig:good-example} above for an example: we give the model
input images $x_0$ and $x_i$, and see realistic results as it peforms three
actions: lifting closing the gripper, picking up the green block, and placing it
on top of the blue block. In the real data set, this action failed because the
robot attempted to place the green block on the red block (next to an obstacle),
but here it makes a different choice and succeeds.
We visualize the $8 \times 8 \times 8$ hidden layer by averaging cross the
channels.

%Our models are capable of generating clear, precise predictions of the future on
%our simulated tasks. Some of these predictions can be seen in
%Fig.~\ref{fig:blocks-example}, earlier in the section. These images tend to be
%very close to the observations of the world from the simulation.

\subsection{Learned Hidden State}

\begin{figure}[bt!]
  \centering
  \includegraphics[width=\columnwidth]{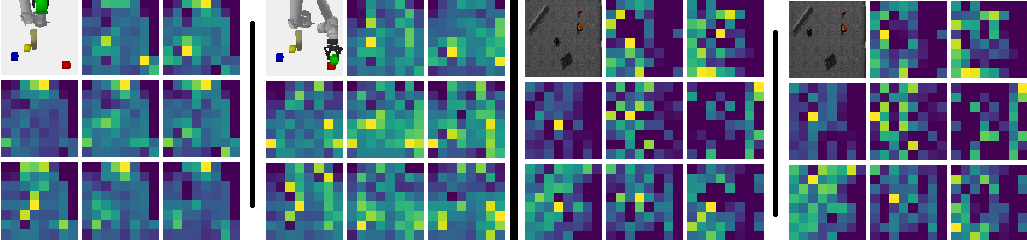}
  \caption{Learned hidden state representations for the simulated stacking task
  (left) and navigation task (right).}
  \label{fig:hidden-states}
\end{figure}

The state learned by our encoder-decoder contains information representing
possible objects and positions.
We compared use of the hidden state representation learned on our data set with
models trained for specific tasks on different models.
Fig.~\ref{fig:hidden-states} shows the lower-dimensional hidden state
representation used by the predictive model.
In the stacking task, we can see how different features change
dramatically as objects are moved about in the scene. In the simpler navigation task, only the Husky
robot changes position. In this case, many of the features are likely redundant.

\begin{figure}[bt!]
  \centering
  \includegraphics[width=\columnwidth]{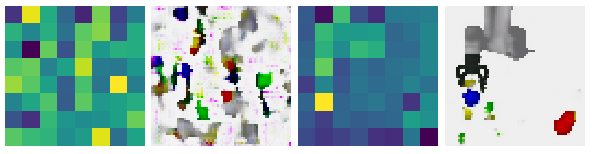}
  \includegraphics[width=\columnwidth]{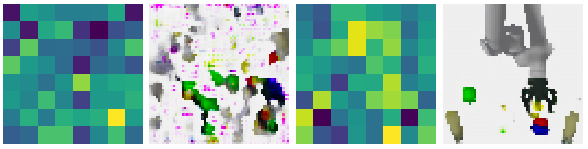}
  \caption{Randomly sampled hidden states projected onto the manifold of the
  transform function $T(h_0,h,a)$.}
  \label{fig:random}
  \vskip -0.5cm
\end{figure}

To visualize the effects of the transform function $T(h_0,h,a)$ on the learned
hidden state, we randomly sampled a number of hidden states and repeatedly
applied $T(h_0,h,a)$ with random actions. After 200 steps, we see results similar to
those in Fig.~\ref{fig:random}, with objects and the arm positioned randomly in
the scene. From left to right, these show (1) the randomly sampled hidden state,
(2) a decoding of this hidden state, (3) a hidden state after 200 random
transform operations, and (4) the decoded version of this hidden state.

\subsection{Model Architecture}\label{sec:results-architecture}

We performed a set of experiments to verify our model architecture, particularly
in comparing different versions of the transform block $T$. We compare three
different options: the block as shown in Fig.~\ref{fig:transform}, the same
block with the skip connection removed, and the same block with the spatial
softmax and dense block replaced by a stride 2 and a stride 1 convolution to
make a more traditional U-net similar to that used in prior
work~\cite{DBLP:journals/corr/IsolaZZE16}.

\begin{figure}[bt!]
	\centering
	\includegraphics[width=\columnwidth]{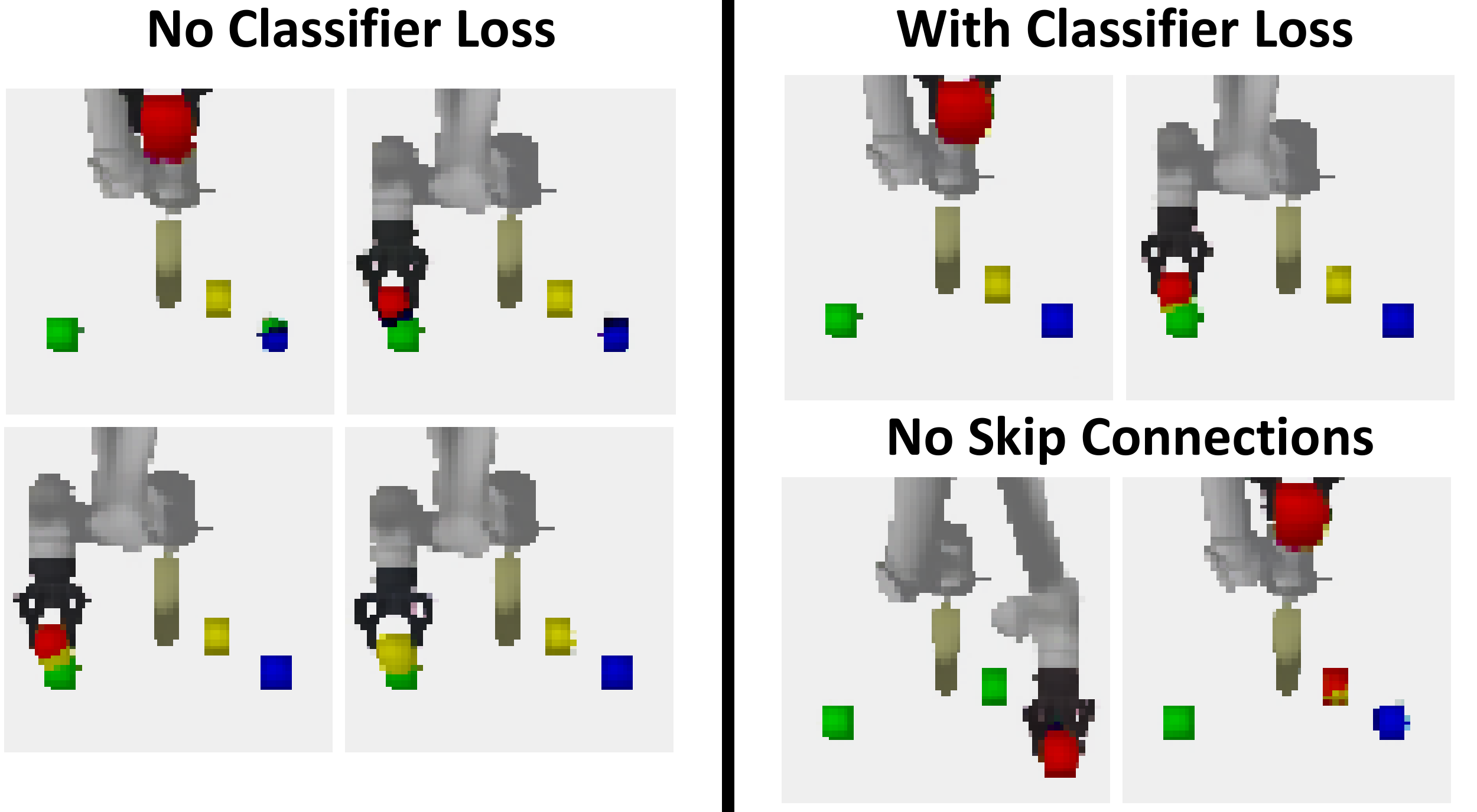}
	\caption{Selected results different possible architectures for the Transform
  block.}
	\label{fig:transform-comparison}
  \vskip -0.5cm
\end{figure}

To compare model architectures and training strategies, we propose a simple
metric: given a single frame, can we determine
which action just occurred? This is computed given the same pretrained
discriminator discussed in Sec.~\ref{sec:learning-approach}.
We compare versions of the loss function with and without the classifier loss
term, and with this term given one of two possible weights.
Both the classifier loss term and the conditional GAN discriminator
term were applied to the second of two transforms, to encourage the model to
generate predictions that remained consistent over time.

\begin{table}[bt]
  \begin{tabular}{c c c c c }
    Model & $x_{1}$ label &  $x_{1}$ error & $x_{2}$ label & $x_{2}$
    error \\
    \hline
    Naive & 87.2\% &0.0161 & 74.3\% & 0.0261 \\
    L1 & 88.1\% & 0.016 & 84.5\% & 0.018 \\
    L1+$0.01C$ & 87.9\% & 0.0177 & 94.3\% & 0.0214\\
    L1+$0.001C$ & 88.2\% & 0.016 & 85.4\% & 0.0184 \\
    No Skips & 87.5\% & 0.0224 & 85.4\% & 0.0247 \\
    cGAN~\cite{DBLP:journals/corr/IsolaZZE16} & 84.5\% & 0.0196 & 77.5\% & 0.0235 \\
  \end{tabular}
  \caption{Comparison of test losses as assessed by image prediction error (MAE) and image confusion on successful examples only.}
  \label{table:losses}
  \vskip -0.5cm
\end{table}

\begin{table}[bt]
  \begin{tabular}{c c c c c }
    Model & $x_{1}$ label &  $x_{1}$ error & $x_{2}$ label & $x_{2}$
    error \\
    \hline
    Naive & 89.3\% &0.0182 & 77.3\% & 0.0276 \\
    L1 & 90.4\% & 0.0181 & 86.4\% & 0.0209 \\
    L1+$0.01C$ & 90.6\% & 0.0198 & 95.6\% & 0.0239\\
    L1+$0.001C$ & 90.9\% & 0.0181 & 88.6\% & 0.0208 \\
    No Skips & 90.1\% & 0.0243 & 89.3\% & 0.0271 \\
    cGAN~\cite{DBLP:journals/corr/IsolaZZE16} & 86.5\% & 0.0216 & 79.9\% &
    0.0260 \\
  \end{tabular}
  \caption{Comparison of test losses as assessed by image prediction error (MAE) and image confusion.}
  \label{table:losses2}
  \vskip -0.5cm
\end{table}

Tables~\ref{table:losses} and \ref{table:losses2} show the results of this comparison.
There were 37453 example frames from successful examples and 54077 total examples in
the data set. In general, the pretrained encoder-decoder structure allowed us to reproduce
high-quality images in all of our tasks.
The ``Naive'' model indicates L1 loss with only one prediction; it performs
notably worse than other models due to errors accumulating over subsequent
applications of $T(h_0,h,a)$.

The  classifier loss term improved the quality of predictions on the
second example, and improved crispness of results, at the slight cost of some
pixel-wise error on the output images.
Here, we see that adding the classifier loss
terms (L1+$0.01C$ and L1+$0.001C$)
slightly improved recognition performance looking forward in time. This
corresponds with increasing image clarity corresponding to the fingers of the
robot's gripper in particular. Pose and texture differences largely explain the
differences in per-pixel error.

The ``No Skips'' model was trained the
same as the L1+$0.001C$ model, but without the skip connections in
Fig.~\ref{fig:transform}. These connections allow us to fill in background
detail correctly (see Fig.~\ref{fig:transform-comparison}), but were not
necessary for the key aspects of any particular action.

The cGAN was able to capture feasible texture, but often
missed or made mistakes on spatial structure. It often misplaced blocks, for
example, or did not hallucinate them at all after a placement action.
This may be because of the noisy data and the large number of failures.

\subsection{Plan Evaluation}

\begin{figure*}[bt!]
	\centering
	\includegraphics[width=2\columnwidth]{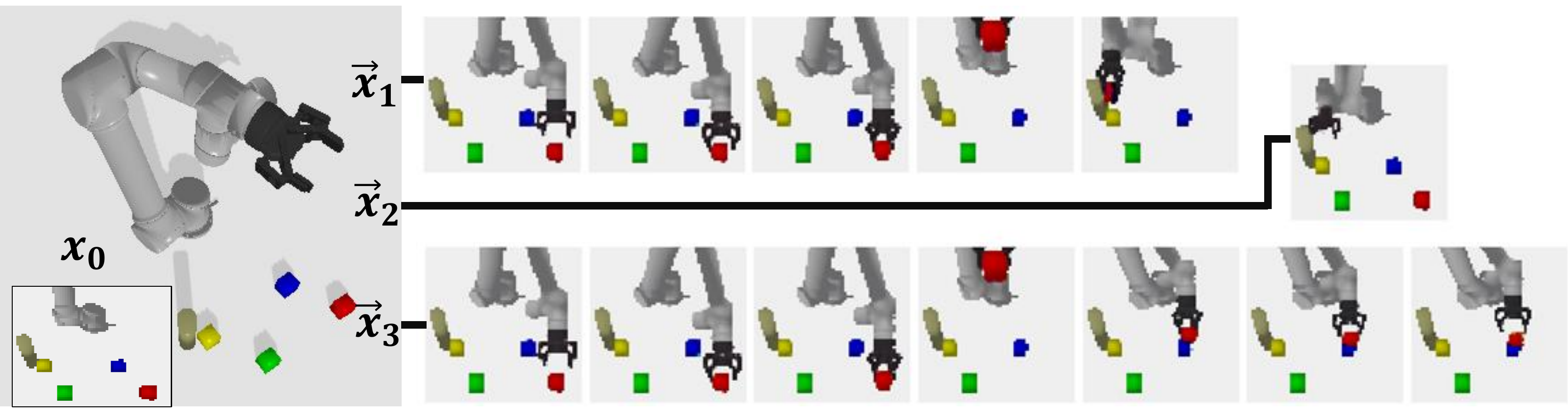}
	\caption{First three results from a call of the planning algorithm on a random
  environment.}
	\label{fig:plans}
  \vskip -0.25cm
\end{figure*}

Our approach is able to generate feasible
action plans in unseen environments and to visualize them; see
Fig.~\ref{fig:plans} for an example. The first two plans are recognized as
failures, and then the algorithm correctly finds that it can pick up the red
block and place it on the blue without any issues. The value function $V(h)$ correctly
identified frames as coming from successful or failed trials $83.9\%$ of the
time after applying two transforms -- good, considering that it is impossible to
differentiate between success and failure from many frames. It correctly
classified possible next actions $96.0\%$ of the time.

\begin{figure}[bt!]
	\centering
	\includegraphics[width=\columnwidth]{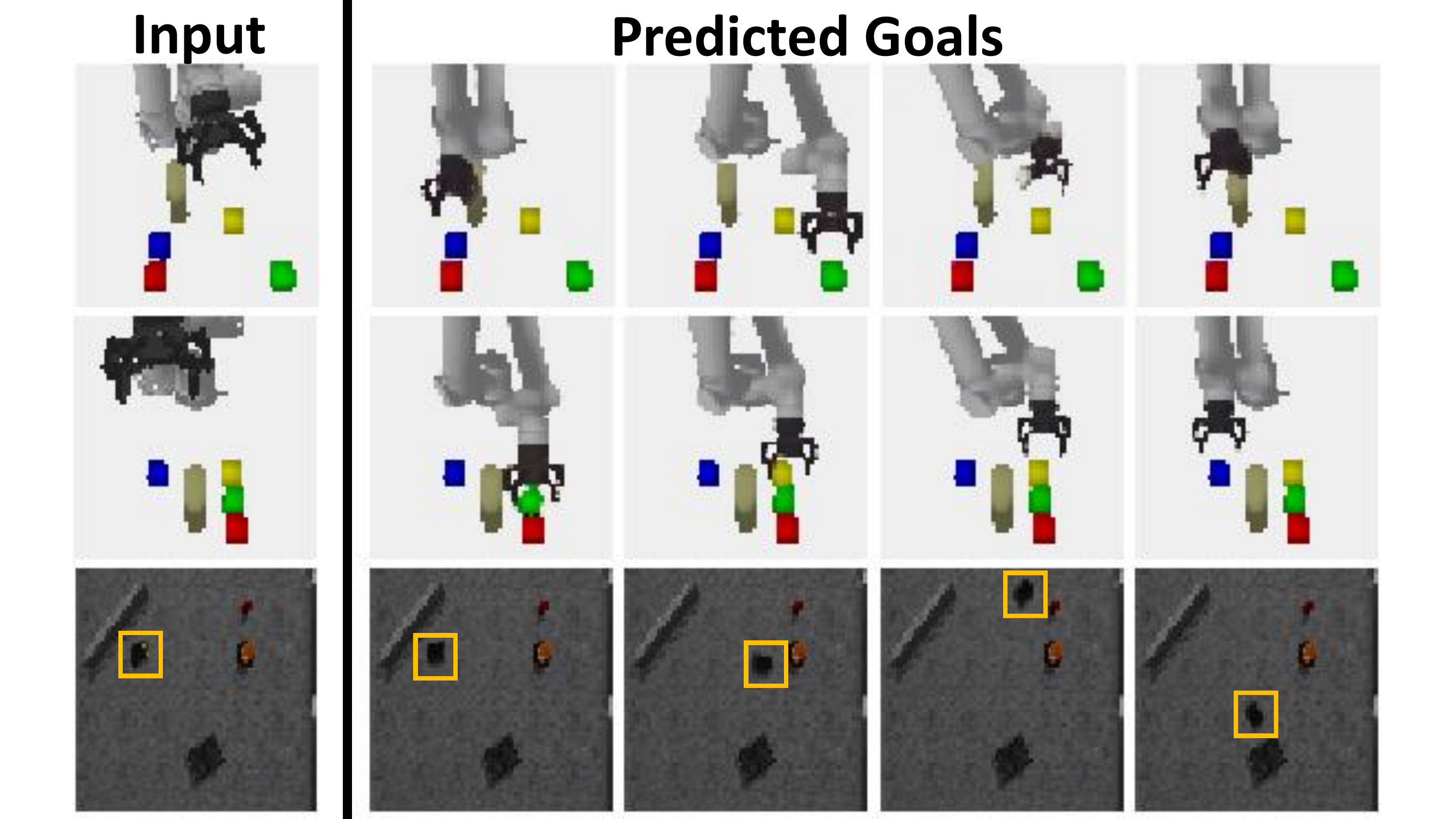}
	\caption{Analyzing parallel possible futures for the stacking and navigation
  tasks. Top row shows multiple good options for grasping separate objects; the
second shows how attempts to grab two objects are clear failures and only one is
a clear success.}
	\label{fig:parallel}
  \vskip -0.5cm
\end{figure}

At each node in our tree search, we examine multiple possible futures. This is
important both for planning and for usability: it allows our system to justify
future results. Fig.~\ref{fig:parallel} shows examples of these predictions in
different environments. We see how the system will predict a set of serious
failures in the middle row, when attempting to grasp the red or blue blocks, and
one possible failure when grasping the yellow block.

We tested our method on $10$ new test environments in the stacking task. On each of these
environments, we performed a search with $10$ samples. Our approach found $8$
solutions to planning tasks executing the demonstrated high-level actions, and
in $2$ tasks it predicted that all of its actions would result in failures,
due to proximity to the obstacle. This highlights an advantage of the visual
task planning approach: in the event of a failure, the robot provides a clear
explanation for why (see the second sample in Fig.~\ref{fig:plans} for an
example).

\subsection{Surgical Image Prediction}

\begin{figure}[bt]
  \centering
  \includegraphics[width=\columnwidth]{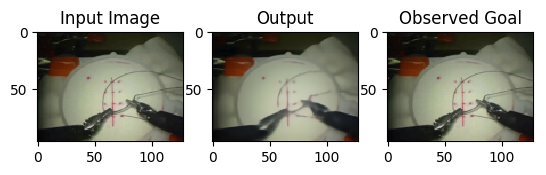}
  \caption{Results of the pretraining step show that our encoder/decoder
    architecture can accurately capture all the relevant detail even in noisy,
  complex scenes.}
  \label{fig:pretrain-suturing}
  \vskip -0.5cm
\end{figure}

We trained our network on 36 examples in the JIGSAWS dataset, leaving out 3
for validation. %In all, the dataset consists of 131,700 frames.
%Due to the size of the dataset, we downsampled the video by a factor of 20.
We were able to generate predictions that clearly showed the location
of the arms after the next gesture, as shown in Fig.~\ref{fig:suturing}.
The learned space $\mathcal{H}$ is very expressive, but loses some fine details such as
the thread at times; see Fig.~\ref{fig:pretrain-suturing}.
The result of $f_{dec}(f_{enc}(x))$ still has almost all the same detail as the goal image (right).

Image prediction created recognizable gestures, such as pulling the
thread after a suture.
While our results are visually impressive, error was higher than
in the robotic manipulation task: we saw mean absolute error of 0.039 and 0.062
for generated images $x_{1}$ and $x_{2}$, respectively.
This is likely because the surgical images contain a lot more subtle but functionally irrelevant data
that is not fully reconstructed by our transform.
It therefore looks ``good enough'' for human perception, but does not compare
as well at a pixel-by-pixel level.
In addition, there is high variability on the performance of each action and a
relatively small amount of avaiable data.
%It's also possible that we're overfitting the data as a result of the need to
%downsample the video (due to memory constraints.)

Again, the cGAN did not have a measurable impact: MAE of 0.039 and 0.067 across
the three test examples.
As such, the longer training time of the cGAN does not seem justified.
In general, it appears to us that conditional GANs are good at modifying
texture, but not necessarily at hallucinating completely new image structure.

\section{Conclusions}\label{sec:conc}

%We described a method for visual task planning with robots from a mixture of
%good and bad human demonstrations.
%This method works
%in multiple domains, and relies on an expressive learned representation trained
%via autoencoder.
We described an architecture for visual task planning, which learns an
expressive representation that can be used to make meaningful predictions
forward in time.
This can be used as part of a planning algorithm that explores multiple
prospective futures in order to select the best possible sequence of future actions to execute.
In the future we will apply our method to real robotic examples and expand
experiments on surgical data.

% can add acknowledgements once it's accepted
\section*{Acknowledgments}
This research project was conducted using computational resources at the Maryland Advanced Research Computing Center (MARCC).
It was funded by NSF grant no. 1637949.
The Titan Xp used in this work was donated by the NVIDIA Corporation.

{
%% Use plainnat to work nicely with natbib. 
\bibliographystyle{IEEEtran}
\small
\bibliography{bib/planners,bib/psych,bib/task_description,bib/taskmodels,bib/drl,bib/software,bib/oac,bib/ltl,bib/representation,bib/prediction,bib/machine_learning,bib/grasping,bib/surgical}
}

\end{document}